\documentclass[runningheads]{llncs2e/llncs}
\usepackage{graphicx}
\usepackage{subfigure}
\usepackage{amsmath} 
\usepackage{amssymb}  
\DeclareMathOperator*{\argmax}{\arg\!\max}

\begin{document}
\title{Probabilistic Graphs for Sensor Data-driven Modelling of Power Systems at Scale}
\titlerunning{Prob. Graphs for Sensor Data-driven Modelling of Power Systems at Scale}
%
\author{Francesco Fusco
}
\authorrunning{F. Fusco}
%
\institute{IBM Research - Ireland
\email{francfus@ie.ibm.com}
}
\maketitle              
\begin{abstract}
The growing complexity of the power grid, driven by increasing share of distributed energy resources and by massive deployment of intelligent internet-connected devices, requires new modelling tools for planning and operation. Physics-based state estimation models currently used for data filtering, prediction and anomaly detection are hard to maintain and adapt to the ever-changing complex dynamics of the power system. A data-driven approach based on probabilistic graphs is proposed, where custom non-linear, localised models of the joint density of subset of system variables can be combined to model arbitrarily large and complex systems. 
The graphical model allows to naturally embed domain knowledge in the form of variables dependency structure or local quantitative relationships. A specific instance where neural-network models are used to represent the local joint densities is proposed, although the methodology generalises to other model classes. Accuracy and scalability are evaluated on a large-scale data set representative of the European transmission grid. 

\end{abstract}

\section{Introduction} \label{sec:introduction}

The increasingly large share of renewable energy generation being injected into the electrical grid, along with the growing diversity and complexity in the mix of distributed resources necessary for balancing supply and demand (e.g. home energy management systems, smart thermostats, electric storage and vehicles) call for novel power system modelling tools. Operational and planning decisions require visibility into a wider range of variables as opposed to the limited view of net power flows and injections available through traditional tools based on first-principle equations, such as power-flow and state-estimation models \cite{Abur2004}. Furthermore, expanding and maintaining such physics-based tools, which involves identifying and updating the right sets of fundamental equations and physical parameters, becomes increasingly harder given the higher degree of complexity of the system dynamics and its fast-changing nature. 

The unprecedented availability of data, resulting from increasing numbers of Internet-of-Things (IOT) sensor devices \cite{Collier2017}, makes data-driven approaches to power system modelling practical. Besides its inherent ability to adapt and grow along with the available data, a data-driven model should serve the key functionalities that are typical of power system state estimation tools \cite{Abur2004}:
\begin{itemize}
\item run model predictions based on assumptions about a subset of variables (short-term grid predictions, what-if scenarios for planning and optimization); 
\item filter observations to reduce sensor noise and reconstruct missing data; 
\item detect and localise system or data anomalies.
\end{itemize}

Numerous classes of machine-learning approaches can be used to model the non-linear power system dynamics and handle data imputation, sensor noise and outliers.  Variational autoencoders provide an efficient approach to denoising \cite{Kingma2014}, although they do not deal naturally with missing data. Non-linear extensions, based on feedforward neural networks, of principal component analysis (NLPCA) \cite{Scholz2005} and factor analysis (NLFA) \cite{Honkela2004} are well suitable to solve the data imputation problem. Non-linear extentions of PCA based on kernel methods \cite{Sanguinetti2006,Nguyen2009} or Gaussian processes \cite{Luttinen2009} can also be used to handle missing data. One limitation of such models, however, is that they cannot naturally account for the factorization and sparsity induced by power systems topology. Even where efficient computational frameworks exist to handle large number of variables, for example in the case of neural networks, the consequent exponential increase in parameters makes their demands in terms of training data not practical in very dynamic and changing sensor data-driven systems.

More recently, computational methods based on probabilistic graphical models and belief propagation were studied to offer a naturally distributed solution to the traditional power systems state-estimation problem \cite{Hu2011,Cosovic2016,Cosovic2016a}. A belief-propagation algorithm supporting a combination of computational nodes representing state-estimation sub-problems, as well as other custom non-linear relations, was proposed in \cite{Fusco2017}. All research on probabilistic graphs applied to power systems, however, focused on representing power-flow equations with a graphical model and on the inference task. While offering improved flexibility, such models still rely on given sets of fundamental equations and knowledge of their physical parameters. Keeping track of the updates in grid topology and parameters is a challenging, time-consuming and at times impossible, task. None of the mentioned research deals with the problem of learning the model relationships fully from the data. 
In this study, a learning algorithm is derived based on the assumption that the non-linear joint densities between subsets of variables are modelled as Gaussian latent-variable models, using a neural-network NLPCA model \cite{Scholz2005}. It is demonstrated how the NLPCA factor nodes naturally handle missing data, thus being well suited for the mentioned power system modelling tasks, which can mostly be reduced to data imputation problems. It is also shown how very simple heuristics based on the power system topology can be used to initialize the graphical model structure such to obtain a more efficient model (in terms of number of parameters) than a centralised NLPCA with a similar modelling accuracy. It is also shown how, in the presence limited training data, the accuracy of the proposed graphical model does not degrade with increasing dimensionality of the problem, in contrast to the centralised approach. 

The model is detailed in Section \ref{sec:methods}. Results on a publicly-available data set representative of the European transmission grid, are reported in Section \ref{sec:results}. Conclusive remarks and scope for further work are outlined in Section \ref{sec:conclusion}. 
\section{Methodology} \label{sec:methods}

Graphical models provide a natural way to represent structural relationships between random variables, to embed domain knowledge and to deal with missing data. The proposed method is derived using the factor graphs representation, which provides a way to link, in the form of conditional and joint probability distributions, potentially very different models together in a principled fashion that respects the rules of probability theory \cite{Frey2005}. 

After introducing the modelling approach and assumptions, in Section \ref{sec:methods_model}, the neural-network NLPCA approach used to model the non-linear joint densities is detailed in section \ref{sec:methods_nlpca}. The inference and learning algorithms are then described in Sections \ref{sec:methods_inference} and \ref{sec:methods_learning}.

\subsection{Model} \label{sec:methods_model}

Given a set of variables, $\mathcal{Z}=\left\{z_1,z_2,\ldots \right\}$, graphical models are a mathematical tool which can be used to express factorizations of the form $\Phi(z_1,z_2,\ldots) = \prod_{j} \phi_j(\mathcal{Z}_j)$, where $\phi_j(\mathcal{Z}_j)$ is a probability density defined on a subset of the variables $\mathcal{Z}_j \in \mathcal{Z}$. In a factor graph parametrization, a \emph{variable node} is defined for each variable $z_i$ and a \emph{factor node} represents the densities $\phi_j$. A graph edge between a factor node and a variable node exists for each $z_i \in \mathcal{Z}_j$. Factor nodes can represent both conditional densities (directed edges) and joint densities (undirected edges), thus offering a very general modelling framework. 

For the specific purposes outlined in section \ref{sec:introduction}, the set of variables $\mathcal{Z}=\left\{\mathcal{X},\mathcal{Y}\right\}$ is a combination of $N$ random state variables $\mathcal{X} = \left\{x_1,x_2,\ldots\,,x_N\right\}$ and  
$M$ sources of noisy observations $\mathcal{Y} = \left\{y_1,y_2,\ldots\,,y_{M}\right\}$. 
The joint probability distribution is factorized as: 
\begin{equation}
p(\mathcal{X},\mathcal{Y}) = \prod_{m=1}^M p(y_m|\mathcal{X}_{m})\prod_{k=1}^K p(\mathcal{X}_k),
\label{eq:density_factorization}
\end{equation}
where $p(y_m|\mathcal{X}_m)$ denotes a conditional probability density, $p(\mathcal{X}_k)$ is a joint probability density and $\mathcal{X}_i \in \mathcal{X}$. It is assumed that both types of relationships can be modelled with Gaussian densities as follows:
\begin{equation}
p(\mathcal{X},\mathcal{Y}) \propto \prod_{m=1}^M e^{-\frac{1}{2}\left[y_m-f_m(\mathcal{X}_m)\right]^\top R_m\left[y_m-f_m(\mathcal{X}_m)\right]} \prod_{k=1}^K e^{-\frac{1}{2}g_k(\mathcal{X}_k)^\top S_k g_k(\mathcal{X}_k)} ,
\label{eq:density_factorization_gaussian}
\end{equation}
where $f_m(\cdot)$, $g_k(\cdot)$ are generally non-linear functions and $R_i$, $S_k$ are covariance matrices. The Gaussian assumption allows for efficient inference and learning algorithms to be designed, as shown in Section \ref{sec:methods_inference} and \ref{sec:methods_learning}, and it is  reasonably accurate when applied to sensor data processing in power systems \cite{Abur1997}.

In (\ref{eq:density_factorization_gaussian}), the conditional densities express explicit relations between the data and the state variables. They can be used to embed domain knowledge into the model such that the state variable has a desired physical meaning. 
Typically, in power systems, the state variables are defined as the set of voltage magnitudes and angles, and all data are related to them through power flow equations parametrized on the topology and impedances of the grid. In a fully data-driven model, however, any other desired representation is possible and, for example, state variables could include, for example, amount of renewable generation, demand- response capacity at the nodes of the grid, as long as an explicit relation to the data exists, as it will be demonstrated in section \ref{sec:results}.

The joint probabilities in (\ref{eq:density_factorization_gaussian}) represent structural relationships between subset of state variables within the model. Domain knowledge and heuristics based on power system topology can give indications on the structure of the joint densities. However, the functional form of the relation, $g(\cdot)$, alon with the covariance $S_k$, is not known in general and should be learned from the data. It is reasonable to assume that the joint density of a set of state variables can be fully specified by a lower-dimensional latent variable, that is $p(\mathcal{X}_k) = p(\mathcal{X}_k | z_k)$. In Section \ref{sec:methods_nlpca}, an approach to model such densities using NLPCA, originally proposed in \cite{Scholz2005}, is described. The methodology, however, is quite general and any other class of latent-variable models, as reviewed in Section \ref{sec:introduction}, could be used at no loss of generality for the inference and learning algorithms derived in Section \ref{sec:methods_inference} and \ref{sec:methods_learning}.

\subsection{Joint factor node based on NLPCA} \label{sec:methods_nlpca}

The NLPCA model considered here is a modification of the classical autoencoder, where only the decoder neural network is learnt from the data, while the encoding function is based on actual model inversion through gradient descent \cite{Scholz2005}, thus providing for a more natural way for handling missing data.  

A feed-forward neural network provides a non-linear mapping of the type: 
\begin{equation}
  \mathcal{X}_k = h_k(z_k,\vartheta_k) + \zeta_k,
\end{equation}
where it is assumed that $\zeta_k \sim \mathcal{N}(0,R_k)$ is a zero-mean Gaussian error term with covariance matrix $R_k$ and $\vartheta_k$ are the network parameters (weights). Note that the input $z_k$ is also unknown and an extension to back-propagation is needed for learning both $\vartheta_k$ and $z_k$ at the same time \cite{Scholz2005}. By including an additional input layer, such that the input to the network is an identity matrix and the weights of the input layer represent the values of $z_k$, the NLPCA model can be conveniently trained with conventional back-propagation \cite{Scholz2005}. 

Inference requires to run back-propagation, with $\vartheta_k$ fixed, in order to estimate the latent variable $z_k$ for a given value of $\mathcal{X}_k$. Such learning, however, is a relatively low-dimensional optimization with respect to the learning stage, and it typically runs very efficiently \cite{Scholz2005}. Missing data are dealt with quite naturally since it suffices to remove the missing components of $\mathcal{X}_k$ from the gradient calculations. 

With respect to the the joint density $p(\mathcal{X}_k) \propto e^{-\frac{1}{2}g_k(\mathcal{X}_k)^\top S_k g_k(\mathcal{X}_k)}$ in (\ref{eq:density_factorization_gaussian}), a factor node based on the NLPCA model is ultimately defined by:
\begin{align}
&g_k(\mathcal{X}_k) = h_k(h_k^{-1}(\mathcal{X}_k)),  \label{eq:nlpca_g}\\
&S_k = (\tilde{H}_k^\top R_k^{-1} \tilde{H}_k)^{-1} \label{eq:nlpca_S}.
\end{align}
As mentioned, the model inversion in (\ref{eq:nlpca_g}), $h^{-1}(\cdot)$, has no explicit form but it represents a gradient-descent procedure. The expression for the covariance matrix in (\ref{eq:nlpca_S}) makes use of a typical approximation from non-linear Gaussian filtering \cite{Abur2004}, where $\tilde{H}_k$ is the Jacobian of the encoding function computed at the solution for the latent variable, namely $\nabla h(z_k)$. The notation $\tilde{H}_k$ is used to highlight the fact that, in the case of partially available data, $\tilde{H}_k$ is only a subset of the full gradient $H_k$, composed of the rows corresponding to the subset of available data. 

The adopted NLPCA model is not a generative model in the sense that no model for $p(z_k)$ is provided, but only an approximation of $p(z_k | \mathcal{X}_k)$. As a consequence, inference can only be solved in the case of partially missing data. While this is generally a limitation, generative models are typically more computationally intensive and, more importantly, there are no practical scenarios for the case of fully missing data. The proposed treatment of probabilistic graphs is, however, not restricted to the particular choice for the latent variable model and generative models, for example based on NLFA \cite{Honkela2004}, can be used instead.
 
\subsection{Inference} \label{sec:methods_inference}

Running inference on graphical models reduces to 
the computation of localised messages between factor and variable nodes, along each edge of the graph defined by the factorization of the density function in (\ref{eq:density_factorization}) \cite{Barber2012}. 
In the particular case of Gaussian joint densities, the process of belief propagation involves Gaussian messages and simplifies to the sum-product algorithm, that is a set of summations and multiplications over the parameters of the Gaussian distributions \cite{Koller2009}.

It is convenient to express the factor densities in the canonical form, $p(x) \propto e^{-\frac{1}{2}x^\top J x + x^\top h}$. Linearization based on a first-order Taylor expansion yields: 
\begin{align}
&J_m = F_m^\top R_m^{-1} F_m \label{eq:J_conditional}\\
&h_m = F_m^\top R_m^{-1} (y_m-f_m(\overline{\mathcal{X}}_m))\label{eq:h_conditional}
\end{align}
for the conditional densities and
\begin{align}
&J_k = G_k^\top S_k^{-1} G_k \label{eq:J_joint}\\
&h_k = G_k^\top S_k^{-1} (\mathcal{X}_k - g_k(\overline{\mathcal{X}}_k))\label{eq:h_joint}
\end{align}
for the joint densities. In (\ref{eq:J_conditional}) to (\ref{eq:h_joint}), $F_m$ and $G_k$ are the Jacobian matrices, respectively, of the functions $f_m(\mathcal{X}_k)$ and $g_k(\mathcal{X}_k)$ computed with respect to a value of the state variable $\overline{\mathcal{X}}_k$. 

As derived in \cite{Fusco2017,Koller2009}, messages from variable $x_i$ to factor $f_j$ are computed as:
\begin{align}
h_{x_i\to f_j} &= \sum_{k \in \mathcal{K}_i\setminus j} h_{f_k\to x_i} \label{eq:msg2f_mu}\\
J_{x_i\to f_j} &= \sum_{k \in \mathcal{K}_i\setminus j}J_{f_k\to x_i}, \label{eq:msg2f_J}
\end{align}
where $\mathcal{K}_i$ is the set of factors $f_i$ connected to the variable $x_i$. Messages from factor $j$ to variable $i$ are calculated based on: 
\begin{align}
h_{f_j\to x_i} &= h_{j} - \sum_{k \in \mathcal{K}_j\setminus i} J_{j}^{jk}(J_{x_k\to f_j} + J_j^{kk})^{-1}(h_{x_k\to f_j}+h_j^{k}) \label{eq:msg2v_mu}\\
J_{f_j\to x_i} &= J_{j} - \sum_{k \in \mathcal{K}_j\setminus i} J_{j}^{jk}(J_{x_k\to f_j} + J_j^{kk})^{-1}J_j^{kj}, \label{eq:msg2v_J}
\end{align}
where $J_j^{jk}$ is the block of the $J_j$ matrix with rows corresponding to the variable $x_j$ and columns 
corresponding to variable $x_k$. Similarly, $h_j^k$ denotes the block of the $h_j$ vector corresponding to the variable $x_j$. 

Under the assumption of tree-structured graph, all messages (\ref{eq:msg2f_mu}) to (\ref{eq:msg2v_J}) can be computed by starting from the leaf factor nodes and
iteratively updating all messages where the required input messages have been processed. If the graph is not a tree, a loopy version of the proposed belief propagation algorithm can be derived to iteratively converge to a solution \cite{Cosovic2016}. 
Once messages from all incoming factors are available, variable estimates can be updated with the following iterative scheme: 
\begin{align}
x_i^{t+1} &= x_i^t + \delta x_i \\
\delta x_i &= \left(\sum_{k \in \mathcal{K}_i} J_{f_k \to x_i}\right)^{-1} \left(\sum_{k \in \mathcal{K}_i} h_{f_k \to x_i}\right),
\label{eq:marginal_update}
\end{align}
with the covariance matrix given by:
\begin{equation}
S_{x_i}^{t+1} = \left(\sum_{k \in \mathcal{K}_i} J_{f_k \to x_i}\right)^{-1}.
\end{equation}

It is interesting to note that, for a tree-structured graph with $n$ variable nodes of dimension $d$, the computational complexity of each iteration (without accounting for the back-propagation in a NLPCA joint factor node) is $o(nd^3)$, as opposed to $o(n^3d^3)$ that would be required by working on the global density directly. Being able to represent the factorization of the problem, which is quite a natural property of physical interconnected, distributed systems such as electrical grids, probabilistic graphs have excellent scalability properties.

\subsection{Learning} \label{sec:methods_learning}

Based on the model definition in Section \ref{sec:methods_model}, it is assumed that sensor data are available for the variables in $\mathcal{Y}$. Given a data set $\mathcal{D}$ of $N$ observations $\left\{y_1^{1:N}, \ldots y_m^{1:N} \right\}$, the graphical model in (\ref{eq:density_factorization_gaussian}) is trained by solving the following maximum-likelihood problem: 
\begin{equation}
\argmax_{\vartheta,\gamma} L(\vartheta, \gamma | \mathcal{D} ) = \prod_{n=1}^N \prod_{m=1}^M p(y_m^n |\mathcal{X}_{m}^n,\vartheta_m)\prod_{k=1}^K p(\mathcal{X}_k^n | z_k^n, \gamma_k),
\label{eq:ml}
\end{equation}
where $\vartheta$ are unknown parameters of the conditional densities and $\gamma$ are parameters of the joint densities, represented as conditional, latent variable models, for example using the NLPCA approach proposed in Section \ref{sec:methods_nlpca}. 

The state variables in $\mathcal{X}$ are latent variables and are never directly observed. Expectation maximization is therefore used to solve (\ref{eq:ml}) by iteratively running an inference to solve for all the latent (or missing) variables and then maximising the likelihood. The Gaussian factorization results in localised likelihood functions so that, given an estimate for the latent variables, the parameters $\vartheta_m$ and $\gamma_k$ can be solved for by running independent maximum likelihood on each individual factor density \cite{Koller2009}. Specifically in the case of the NLPCA models, a modified gradient descent minimising the squared error is applied, as described in Section \ref{sec:methods_nlpca}. Similarly, an independent estimator for each parametric conditional density can be designed. In the present study, the conditional densities are assumed to be known as simple identity mapping from the observations to the latent variables, such that  $\mathcal{X}$ assumes a well-defined physical meaning. Learning of the conditional densities is therefore not considered here. 
\section{Results} \label{sec:results}


\subsection{Data preparation and model design} \label{sec:results_data}

\begin{figure}
\centering
\subfigure[]{
\label{fig:dataset_topology}
\includegraphics[width=0.53\textwidth]{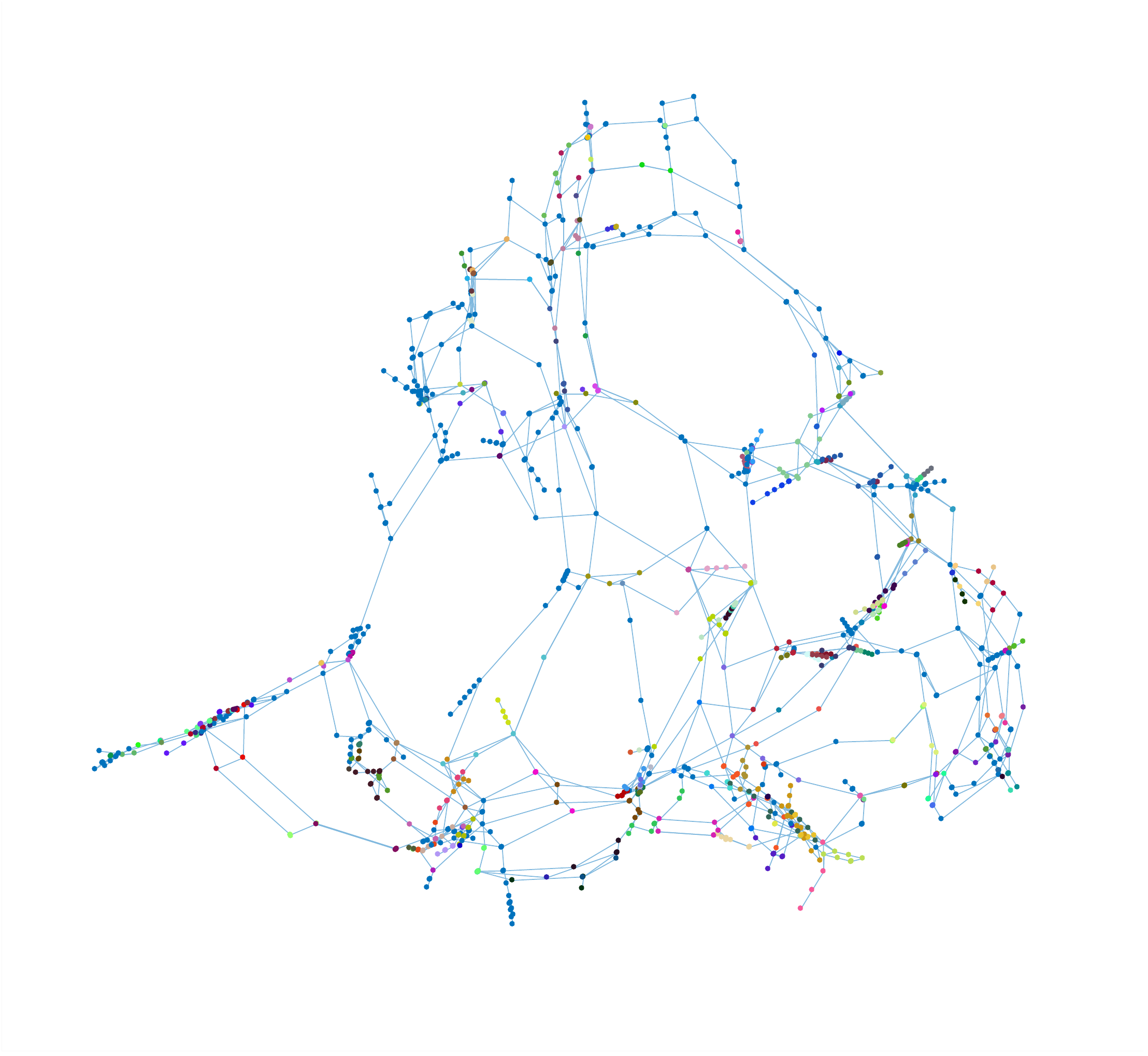}}
\subfigure[]{
\label{fig:dataset_timeseries}
\includegraphics[width=0.44\textwidth]{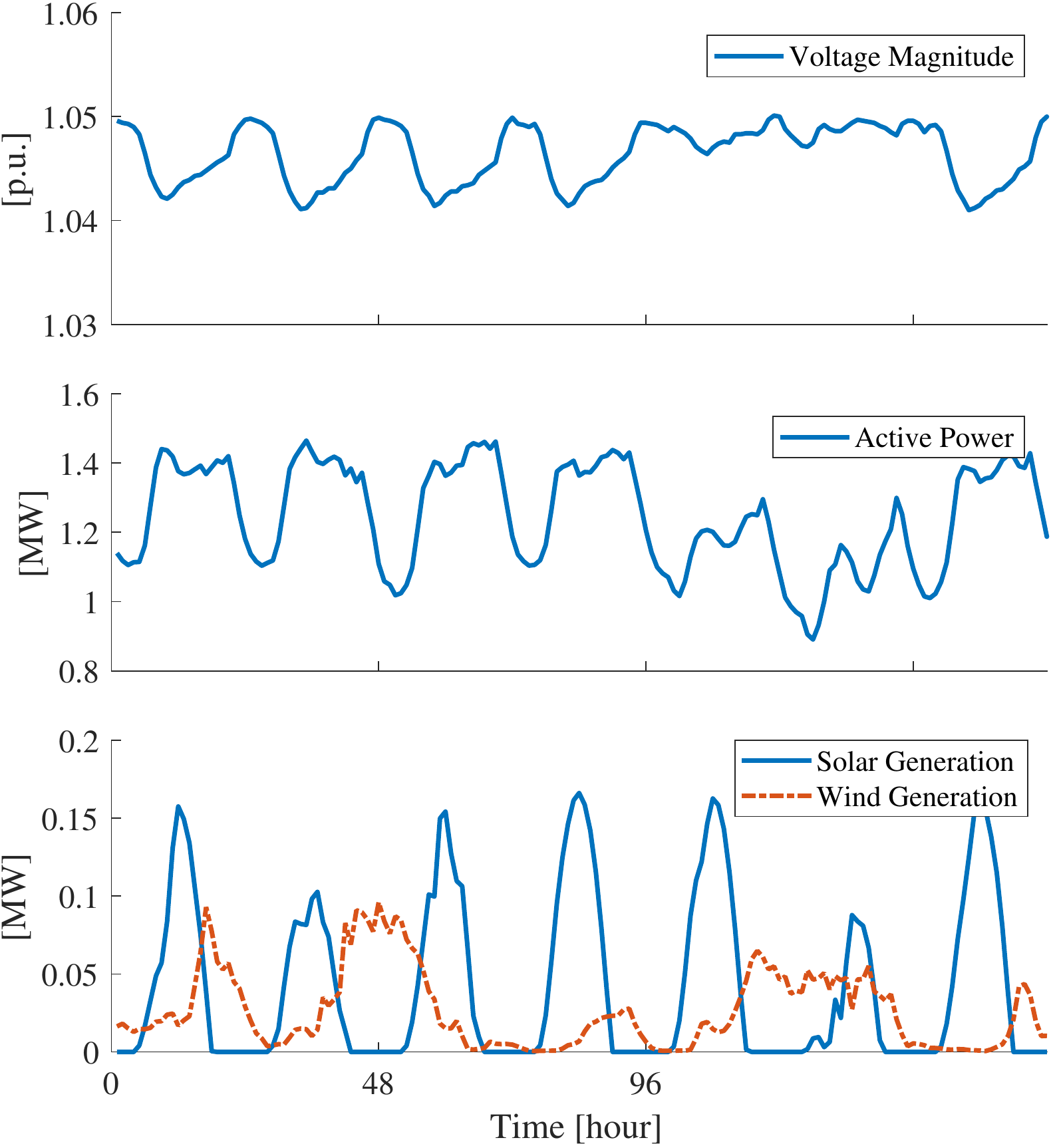}}
\caption{(a) Connectivity of the 1354-bus power grid. The color-coding represents the 273 graph partitions obtained using the spectral method based on Fiedler eigenvectors, as described in section \ref{sec:results_data}. (b) Sample data of voltage magnitude, active power, solar and wind generation at a grid location.}
\label{fig:dataset}
\end{figure}

A publicly available large-scale dataset, created for modelling electricity demand and renewable generation in the European transmission grid, was used \cite{Jensen2017}. The data set includes hourly time series data of energy demand, solar generation and wind generation at nearly 1500 points across Europe from the year 2012 through to 2014. The data set was augmented by running a power-flow simulation to generate time series of voltage and reactive power injections, using the 1354-bus electrical model of the European transmission grid made available in \cite{Josz2016} and the Matpower software \cite{Zimmerman2011}. The bus active power injections where generated by combining loads, wind and solar generation signals at random locations of the data in \cite{Jensen2017}, while the reactive power injections were based on a 0.9 power factor. Sensor noise was simulated as zero-mean, Gaussian random error with standard deviation of $1e^{-3}$ for power measurements and of $1e^{-5}$ for voltage measurements. As a result, a set of 8124 hourly time series data of active/reactive power, voltage magnitude, energy demand, solar and wind generation at 1354 locations was available, for 3 years. Figure \ref{fig:dataset} shows the grid topology and sample time-series data at one grid location. 

\begin{figure}
  \centering
  \includegraphics[width=\linewidth]{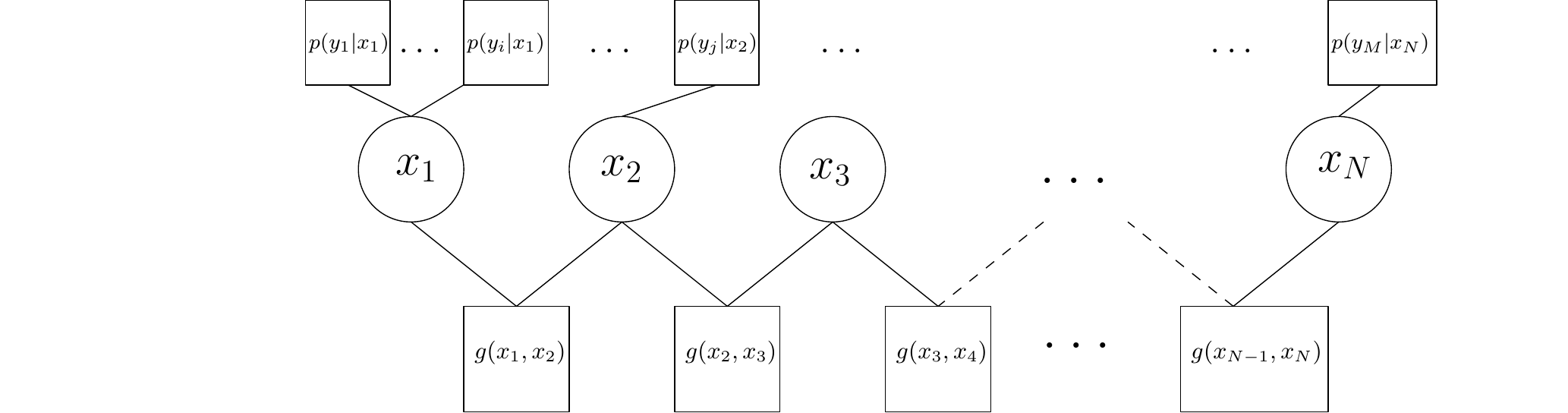}
  \caption{Topology of the proposed graphical model.}
  \label{fig:model_topology}
\end{figure}

The factorization of the proposed graphical model was defined from the grid connectivity based on simple heuristics. The network graph is partitioned using the laplacian of the connectivity matrix and the associated Fiedler eigenvectors \cite{Fiedler1989}, by iteratively bi-sectioning according to the eigenvector corresponding to the second smallest eigenvalue of the Laplacian matrix. The color-coding in Fig. \ref{fig:dataset_topology}  indicates the partitioning obtained after 9 iterations, which resulted in 273 sections with a number of nodes ranging from 1 to 28. In relation to the model described in section \ref{sec:methods_model}, a variable node representing the set of electrical quantities (voltage, active/reactive power, demand, solar/wind generation) at the nodes of each partition was defined. A conditional factor for each metered time series was defined as a simple identity mapping, with a covariance matrix based on the assumed sensor noise. A NLPCA-based joint factor node was introduced for each pair of variable nodes of directly connected graph partitions. The NLPCA model of each joint factor node was based on a 3-layered, fully-connected, neural network with a linear output layer and a sigmoid activation in the hidden layer. The dimensionality of the latent variable was set to half the dimensionality of the output. Loops in the graph were avoided by only connecting variables to at most two joint factors. The final topology of the proposed graphical model is shown in Fig. \ref{fig:model_topology}. Note that many other possible factorizations and corresponding topological structures could be defined, based on alternative heuristics or data-driven methods. Such investigation is not in scope of the current study and the proposed structure, named as Graph-NLPCA, compares favourably with a centralised model using NLPCA, as discussed in the Sections \ref{sec:results_validation}-\ref{sec:results_scalability}. 

As discussed in Section \ref{sec:methods_learning}, model training is based on a few iterations of inference on the factor graph and learning of the individual NLPCA models. Given the flat structure of the model, it was found that less than 5 iterations are sufficient to converge to an accurate model. Back-propagation training and inference on the NLPCA models was performed using the TensorFlow computational library, using mean square error objective function and stochastic gradient descent \cite{GoogleBrain2016}. Data for the month of July 2014 were used for training, and validation was done on the month of August 2014. This is to reflect typical applications where the power system dynamics change over time, due to topological changes (grid expansion, reinforcements) and power flow variations (new customer connections, new renewable generation installations), and it is important that the models are updated on a regular basis only on a short history of the data. Note that while electrical load and generation time series typically express strong seasonality, that is only expected to play a limited role, given the spatial nature of the model. In particular, the proposed model is aimed at capturing the relationship between grid quantities at different locations, rather than the dynamics of a given electrical quantity at different points in time.

\subsection{Model accuracy on missing data} \label{sec:results_validation}

As outlined in Section \ref{sec:introduction}, the primary objective of a data-driven power system model is to being able to handle missing data, such to solve both as a prediction model and sensor data filtering functions. A range of experiments was designed to evaluate the estimation error of the proposed Graph-NLPCA under different proportions of missing data. As a baseline, a centralised approach where all the variables are included in a single NLPCA model is considered. Only part of the complete transmission grid model in the available data is considered, made of 10 grid locations for a total of 375 variables. For this particular dimensionality of the problem, the number of parameters of a NLPCA model, 211500, is comparable with the Graph-NLPCA, 96078.

\begin{table}
\centering
\setlength\tabcolsep{0.3cm}
\begin{tabular}{|c||c|c|c|c|c|}
\hline
 Missing data ratio & $0.1$ & $0.2$ & $0.3$ & $0.4$ & $0.5$ \\
 \hline
\hline
 Graph-NLPCA & $0.0026$ & $0.0040$ & $0.0051$ & $0.0061$ & $0.0077$ \\
 NLPCA & $0.0043$ & $0.0043$ & $0.0045$ & $0.0048$ & $0.0071$ \\
\hline
\end{tabular}
\caption{Root-mean-square error (RMSE) on the estimates for different proportions of missing data. }
\label{tab:missing_data}
\end{table}

\begin{figure}
  \centering
  \includegraphics[width=\linewidth]{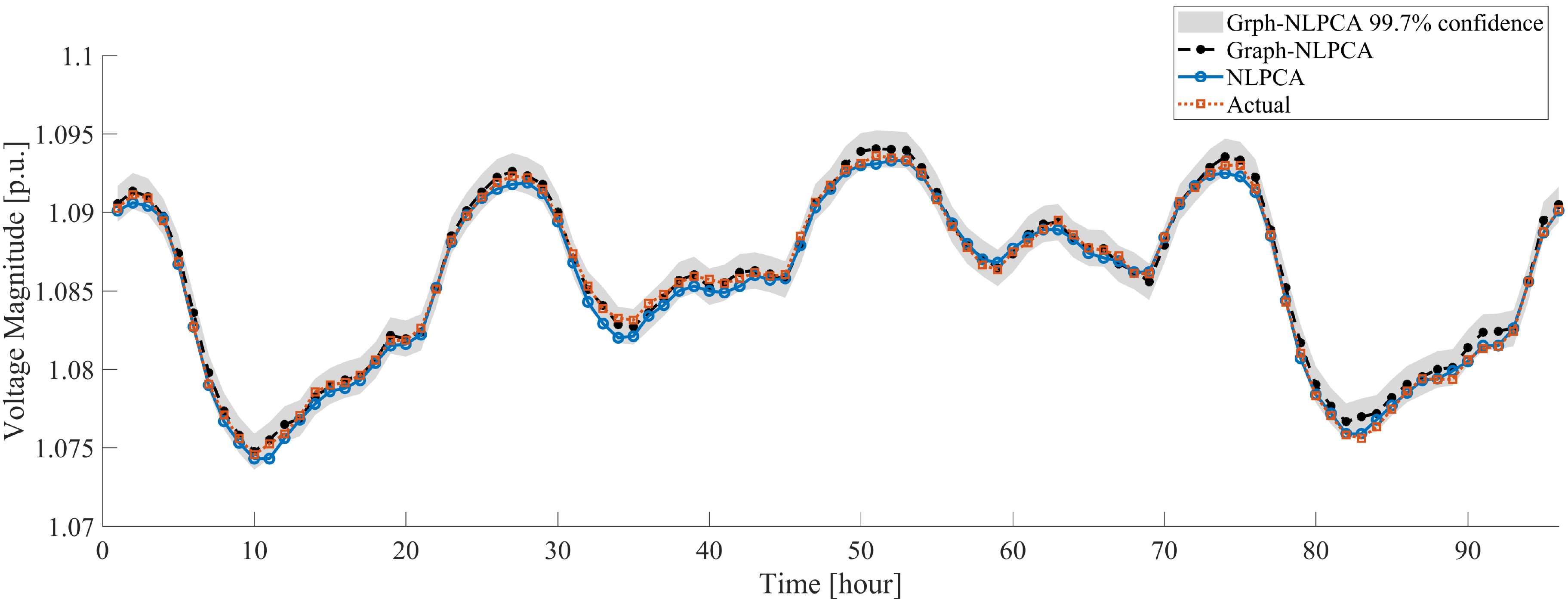}
  \caption{Example of voltage estimation given only data of power and renewable generation.}
  \label{fig:missing_voltages}
\end{figure}

Table \ref{tab:missing_data} summarised the root-mean-square error (RMSE) obtained for a proportion of 10\% to 50\% of missing data. Based on the results, it can be concluded that the factorization enforced by the proposed Graph-NLPCA model does not affect negatively the modelling accuracy with respect to the case where no assumptions about variable dependency is made.

Figure \ref{fig:missing_voltages} shows example results where all the node voltages are estimated given only the power injections, which is a typical use-case for power-flow simulation models. Again, no noticeable difference between the Graph-NLPCA and NLPCA model is observed. 

\subsection{Detecting erroneous data} \label{sec:results_erroneous}

\begin{figure}
  \centering
  \includegraphics[width=\linewidth]{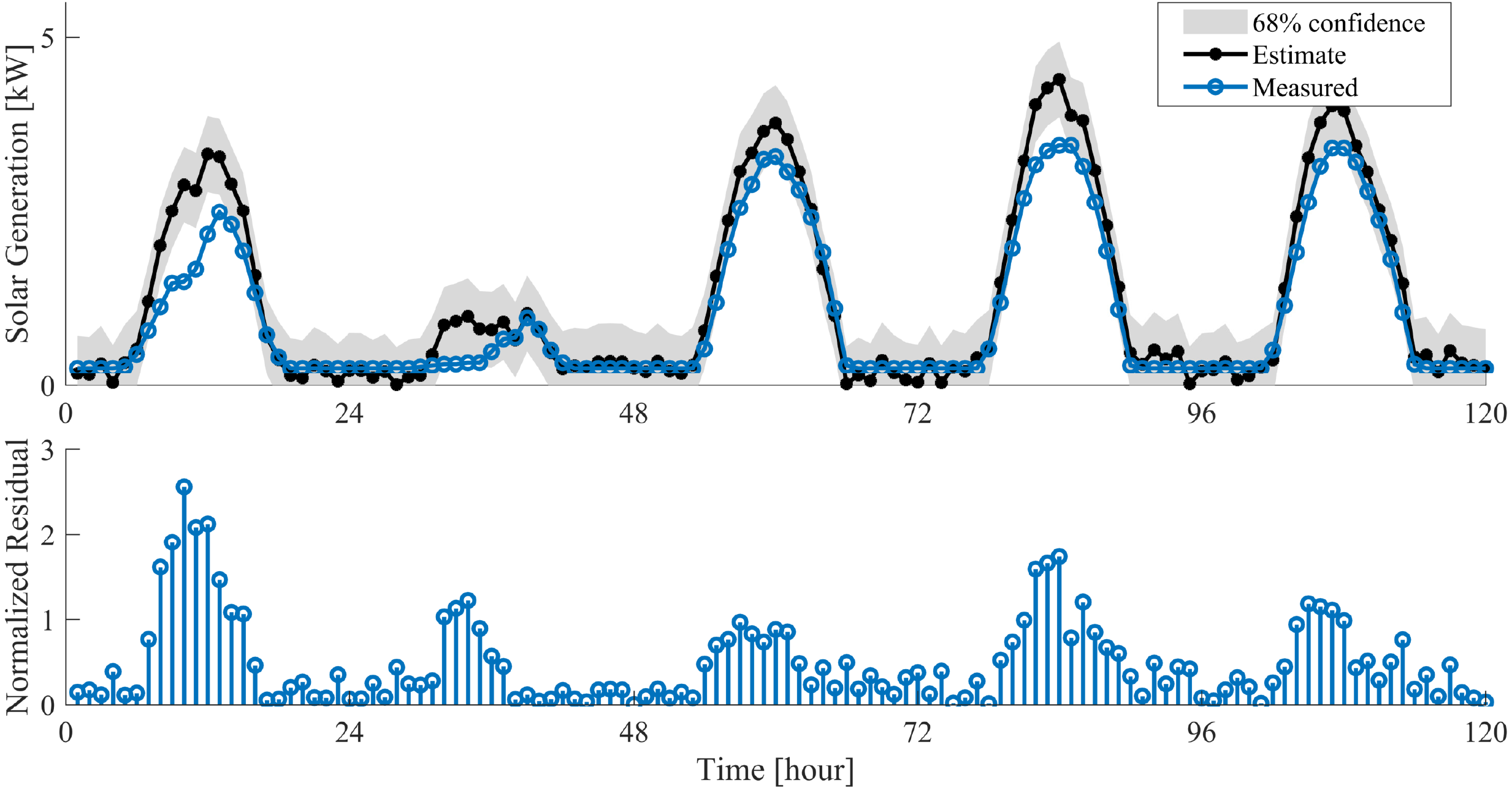}
  \caption{Example of detection of unaccounted solar generation.}
  \label{fig:bad_solar}
\end{figure}

As outlined in section \ref{sec:introduction}, a desired feature of a data-driven power system model is the ability to perform anomaly detection, also referred to as bad data analysis \cite{Abur2004}. A common example is the installation of new renewable generation plant, which may come on-line well before any metering data are made available. 

An experiment was designed by modifying the validation data based on a power-flow simulation after the solar generation data for one of the points was doubled. The measurements for the particular solar generation point, however, was kept at the same level as in the training data. Figure \ref{fig:bad_solar} shows how the resulting estimate is consistently higher than the measurement and a statistical test can be designed to detect the anomaly (a Z-test of the residuals, with respect to the standard deviation produced by the model, gives a probability of about 0.9999 that the estimate is larger than the data).

\subsection{Scalability} \label{sec:results_scalability}

In order to demonstrate the excellent scalability properties of the proposed graphical model, a series of experiments was run where the dimensionality of the problem was increased from the 10 graph partitions (375 variables) used for validation in Section \ref{sec:results_validation}, to 50 (1416), 100 (2576), 150 (3719) and 200 (4643). 

\begin{figure}
  \centering
  \includegraphics[width=\linewidth]{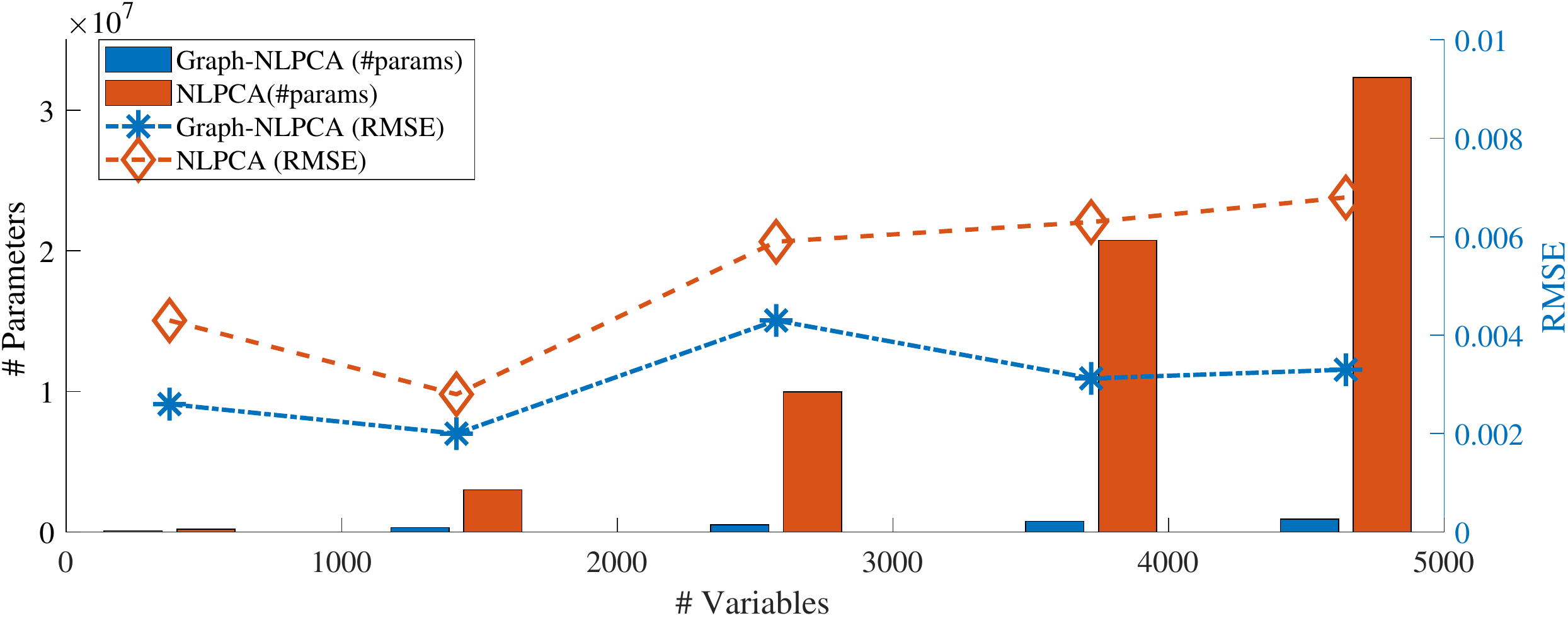}
  \caption{Scalability of number of parameters and modelling accuracy with respect to variable dimensionality. Accuracy is based on the estimation error when 10\% of the data are missing. }
  \label{fig:scalability}
\end{figure}

Figure \ref{fig:scalability} compares the RMSE estimation accuracy obtained with Graph-NLPCA and NLPCA when 10\% of the data was missing on the validation set. While the two models behave similarly up to around 2000 variables, the performance of the centralised approach based on NLPCA starts deteriorating thereafter as the dimensionality of the problem increases. This is to be expected as the number of parameters, also shown in Fig. \ref{fig:scalability}, grows exponentially thus negatively affecting the training phase where only limited data are available. The proposed Graph-NLPCA, on the other hand, maintains a consistent level of error accuracy throughout. 

\section{Conclusion} \label{sec:conclusion}

An innovative approach to data-driven power system modelling based on probabilistic graphs was introduced. By providing a flexible computational framework where custom
non-linear, localised models of the joint density can be combined to model arbitrarily large and complex systems, the method is well suited to handle the complexity and scale of the electrical grid. 

A specific instance of the graphical model, where the joint densities are modelled with a neural-network approach to NLPCA, was evaluated on data imputation and anomaly detection applied to a publicly available data set representative of the European transmission grid. The graph structure is based on a simple heuristics applied to the grid topology and allows for excellent scalability properties, where the model accuracy is not negatively affected by growing dimensionality when only limited training data are available, as opposed to a centralised model. 

Many other possibilities exist to model the individual factor nodes and different models could even be combined in the same graph. Identifying the most suitable model (or set of models) in the context of power systems will be an important topic for future investigation. A naive approach was considered to initialize the structure of the graphical model and applying more formal data-driven techniques will also be scope for further studies. Finally, expanding the model to include time dependencies, such to cover a wider range of prediction tasks and to increase robustness to missing data, will also be an interesting research direction. 

\section*{Acknowledgements} 
This research has received funding from the European Research Council under the European Union’s Horizon 2020 research and innovation programme (grant agreement no. 731232). The author would like to thank Sean McKenna and Bradley Eck, from IBM Research Ireland, for pointing to the data set used in this research, and Michele Berlingherio, from IBM Research Ireland, for providing excellent feedback on the manuscript. 

%
%
%
\bibliographystyle{llncs2e/splncs04}
\bibliography{Energy}

\begin{thebibliography}{10}
\providecommand{\url}[1]{\texttt{#1}}
\providecommand{\urlprefix}{URL }
\providecommand{\doi}[1]{https://doi.org/#1}

\bibitem{GoogleBrain2016}
Abadi, M., {\it et al}: Tensorflow: A system for large-scale machine learning.
  Proceedings of the 12th USENIX Symposium on Operating Systems Design and
  Implementation (OSDI)  (2016)

\bibitem{Abur2004}
Abur, A., Exposito, G., A.: {Power System State Estimation: Theory and
  Implementation}. CRC Press (2004)

\bibitem{Barber2012}
Barber, D.: Bayesian Reasoning and Machine Learning. Cambridge University Press
  (2012)

\bibitem{Collier2017}
Collier, S.E.: The emerging enernet: Convergence of the smart grid with the
  internet of things. IEEE Industry Appl. Magazine  \textbf{23}(2) (2017)

\bibitem{Cosovic2016a}
Cosovic, M., Vukobratovic, D.: Distributed gauss-newton method for ac state
  estimation: A belief propagation approach. IEEE Int. Conf. on Smart Grid
  Comm.  (2016)

\bibitem{Cosovic2016}
Cosovic, M., Vukobratovic, D.: State estimation in electric power systems using
  belief propagation: An extended dc model. IEEE International workshop on
  Signal Processing advances in Wireless Communications (SPAWC)  (2016)

\bibitem{Fiedler1989}
Fiedler, M.: Laplacian of graphs and algebraic connectivity" and combinatorics
  and graph theory. Banach Center Publications  \textbf{25}(1),  57--70 (1989)

\bibitem{Frey2005}
Frey, B.J., Jojic, N.: A comparison of algorithms for inference and learning in
  probabilistic graphical models. IEEE Transactions on Pattern Analysis and
  Machine Intelligence  \textbf{27}(9) (2005)

\bibitem{Fusco2017}
Fusco, F., Thirupathi, S., Gormally, R.: Power systems data fusion based on
  belief propagation. Proceedings of the IEEE PES Innovative Smart Grid
  Technologies (ISGT) Conference Europe  (2017)

\bibitem{Honkela2004}
Honkela, A., Valpola, H.: Unsupervised variational bayesian learning of
  nonlinear models. Proceedings of the Advances in Neural Information
  Processing Systems (NIPS) 17  (2004)

\bibitem{Hu2011}
Hu, Y., Kuh, A., Kavcic, A., Yang, T.: {A Belief Propagation Based Power
  Distribution System State Estimator}. IEEE Computational Intelligence
  Magazine (AUGUST),  36--46 (2011)

\bibitem{Jensen2017}
Jensen, T.V., Pinson, P.: Re-europe and a large-scale dataset for modeling a
  highly renewable european electricity system. Scientific Data 4:170175
  (2017)

\bibitem{Josz2016}
Josz, C., Fliscounakis, S., Maeght, J., Panciatici, P.: Ac power flow data in
  matpower and qcqp format: itesla and rte snapshots and and pegase.
  arXiv:1603.01533  (2016)

\bibitem{Kingma2014}
Kingma, D.P., Welling, M.: Auto-encoding variational bayes. Proceedings of the
  Sixth International Conference on Learning Representations (ICLR)  (2014)

\bibitem{Koller2009}
Koller, D., Friedman, N.: Probabilistic Graphical Models. MIT Press (2009)

\bibitem{Luttinen2009}
Luttinen, J., Llin, A.: Variational gaussian-process factor analysis for
  modeling spatio-temporal data. Advances in Neural Information Processing
  Systems (NIPS) 22  (2009)

\bibitem{Nguyen2009}
Nguyen, M.H., {De la Torre}, F.: Robust kernel principal component analysis.
  Advances in Neural Information Processing Systems (NIPS) 22  (2009)

\bibitem{Sanguinetti2006}
Sanguinetti, G., Lawrence, N.D.: Missing data in kernel pca. Proceedings of the
  European Conference on Machine Learning pp. 751--758 (2006)

\bibitem{Scholz2005}
Scholz, Matthias, Kaplan, Fatma, Guy, L, C., Kopka, Joachim, Selbig, Joachim:
  {Gene expression Non-linear PCA : a missing data approach}. Bioinformatics
  \textbf{21}(20),  3887--3895 (2005). \doi{10.1093/bioinformatics/bti634}

\bibitem{Zimmerman2011}
Zimmerman, R.D., Murillo-Sánchez, C.E., Thomas, R.J.: Matpower: Steady-state
  operations and planning and analysis tools for power systems research and
  education. Power Systems and IEEE Transactions on  \textbf{26}(1),  12--19
  (2011)

\end{thebibliography}

\end{document}